# nnFilterMatch: A Unified Semi-Supervised Learning Framework with Uncertainty-Aware Pseudo-Label Filtering for Efficient Medical Segmentation

Yi Yang

***Abstract*—Semi-supervised learning (SSL) has emerged as a promising paradigm in medical image segmentation, offering competitive performance while substantially reducing the need for extensive manual annotation. When combined with active learning (AL), these strategies further minimize annotation burden by selectively incorporating the most informative samples. However, conventional SSL_AL hybrid approaches often rely on iterative and loop-based retraining cycles after each annotation round, incurring significant computational overhead and limiting scalability in clinical applications. In this study, we present a novel, annotation-efficient, and self-adaptive deep segmentation framework that integrates SSL with entropy-based pseudo-label filtering (FilterMatch), an AL-inspired mechanism, within the single-pass nnU-Net training segmentation framework (nnFilterMatch). By selectively excluding high-confidence pseudo-labels during training, our method circumvents the need for retraining loops while preserving the benefits of uncertainty-guided learning. We validate the proposed framework across multiple clinical segmentation benchmarks and demonstrate that it achieves performance comparable to or exceeding fully supervised models, even with only 5%–20% labeled data. This work introduces a scalable, end-to-end learning strategy for reducing annotation demands in medical image segmentation without compromising accuracy. Code is available here: https://github.com/Ordi117/nnFilterMatch.git.

## I. INTRODUCTION

Medical image segmentation plays an important role in clinical diagnosis and treatment planning by enabling precise delineation of anatomical structures [1]. With the progress of deep learning, segmentation models have demonstrated strong performance across tasks [2]. However, most of these models rely on fully supervised learning (SL) and require large volumes of annotated training data. In clinical field, obtaining pixel-wise annotations is both expensive and time-consuming, often requiring expert radiologists [3,4]. For example, manually annotating more than 8,000 3D CT images for organ segmentation might cumulatively require over 30 years of expert effort.

To reduce reliance on exhaustive labeling, semi-supervised learning (SSL) methods have emerged, leveraging both labeled and unlabeled data. Pseudo-labeling, as introduced by Lee [5], shows promise in reducing test error on MNIST using only 600 labeled samples. However, this method is vulnerable to low-confidence predictions, potentially reinforcing errors during training [6,7]. Alternatively, consistency regularization enforces stable predictions under input perturbations [8,9]. The Mean Teacher model [10] is a representative approach that improves segmentation performance through student and teacher model consistency, as shown in brain lesion segmentation tasks [11,12]. However, it introduces additional complexity by requiring multiple models.

Recent single-model frameworks, such as FixMatch [13] and FlexMatch [14], combine pseudo-labeling with strong-weak augmentation consistency and confidence calibration, respectively. However, they typically treat all unlabeled samples equally, which may limit learning efficiency under complex medical scenarios.

Active learning (AL) offers a complementary strategy by selecting informative unlabeled samples for annotation based on uncertainty or diversity measures [15]. Hybrid SSL_AL frameworks aim to combine pseudo-labeling with sample querying, but they often rely on multi-stage pipelines with expensive retraining [16–18]. Additionally, most approaches use fixed backbone architectures like U-Net or V-Net, which may not generalize well across modalities and require manual tuning.

To address these limitations, we propose a fully automated, annotation-efficient framework that unifies SSL with AL-inspired uncertainty filtering. Built upon the nnU-Net [19], our method automatically adapts preprocessing, architecture, and postprocessing to diverse medical segmentation tasks. We integrate FixMatch-style training, using weakly augmented pseudo-labels as supervision for strongly augmented views. These pseudo-labeled samples are jointly trained with labeled data in a unified pipeline.

Inspired by active learning, we incorporate an entropy-based filtering mechanism that selectively retains uncertain pseudo-labeled samples during training. In contrast to conventional active learning approaches that rely on iterative human annotation and retraining cycles, our method performs

Yi Yang is with School of Engineering, Purdue University, West Lafayette, IN 47906, USA. (e-mail: yiy888268@gmail.com).

Yi Yang Contribute to this Paper.

uncertainty-guided sample selection in a fully automated manner, without additional supervision. By emphasizing ambiguous and challenging regions while discarding overconfident predictions, this strategy enhances the model's generalization and improves segmentation performance, particularly in low-annotation regimes.

On the ACDC cardiac segmentation benchmark, our method achieves Dice scores of 89.56% with 5% labeled data and 90.06% with 10%, outperforming existing SSL baselines. The framework is extendable to other tasks such as brain tumor and organ-at-risk segmentation and requires minimal manual tuning thanks to its integration with nnU-Net.

The main contributions of this work are:
1. We propose a fully automated, end-to-end annotation-efficient segmentation framework that integrates semi-supervised learning and active learning-inspired entropy filtering within the nnU-Net framework.
2. We introduce an uncertainty-based filtering strategy for unlabeled data, eliminating the need for iterative annotation while improving training robustness.

We demonstrate state-of-the-art (SOTA) performance on the ACDC dataset under low-label regimes, showing effectiveness of our method in real-world segmentation scenarios.

## II. Related Work

### A. Segmentation Model for Medical Imaging

Recent advances in medical image segmentation have explored architectures beyond convolutional models, including Transformer-based networks and emerging Mamba-based architectures [20,21]. These models offer strong global context modeling and long-range dependency capture, which enhances segmentation performance. While these models have shown promising results, they often require extensive dataset-specific preprocessing and careful fine-tuning of model structures [22]. This reliance on strong technical expertise and manual adjustment makes them inefficient and impractical for deployment in typical medical workflows, where clinical teams may lack the resources or expertise to optimize deep learning frameworks.

To address these challenges, nnU-Net, originally published in 2021 and further refined in recent versions, offers a self-configuring segmentation framework that dynamically adapts its architecture, preprocessing, training schedule, and postprocessing to the specific characteristics of the input dataset. Its adaptive framework, covering everything from resampling and normalization to model depth and patch size, eliminates the need for manual hyperparameter tuning or domain-specific customization. This enables rapid and robust deployment across a wide range of medical imaging tasks and modalities, from MRI and CT to ultrasound, with minimal expert intervention. By integrating SSL and AL-inspired techniques directly into the nnU-Net framework, we aim to create a framework that is not only performant but also annotation- and deployment-efficient, bringing automated segmentation closer to practical clinical use.

### B. Semi-Supervised Learning in Medical Image Segmentation

Obtaining pixel-level annotations for medical images is time-consuming and labor-intensive. SSL seeks to alleviate this burden by utilizing both labeled and unlabeled data during training. One common SSL strategy is pseudo-labeling, where a model generates artificial labels for unlabeled data based on its own confident predictions [9]. Although effective in some cases, pseudo-labeling can lead to confirmation bias and error propagation, particularly in the early training stages.

Consistency regularization has been proposed to mitigate this issue by enforcing that model predictions remain stable under different input perturbations. One representative framework is the Mean Teacher model [11], which trains a student network to match the predictions of a slow updated teacher model. Perone *et al.* [23] successfully applied this framework to multi-center MRI segmentation, demonstrating that consistency-based SSL methods are promising in low-annotation regimes.

Subsequent advances have combined consistency with stronger data augmentations. MixMatch [24] and UDA [25] integrate techniques like entropy minimization and augmentation-aware regularization to improve robustness. Among these, FixMatch [13] stands out for its simplicity and strong empirical performance. It combines pseudo-labeling with consistency enforcement in a single-model framework by applying confident predictions on weakly augmented samples as supervision for strongly augmented versions. Segmentation-specific adaptations such as CrossMatch [26] extend these ideas through dual encoders and self-distillation to improve stability. Inspired by these advances, our work builds on this line of research by incorporating the core idea of weak-to-strong consistency with pseudo-labeling into the nnU-Net [19] backbone. Unlike previous studies, we focus on improving sample efficiency through selective pseudo-label filtering and hybrid learning strategies.

### C. Hybrid Semi-Supervised Learning

While SSL improves data efficiency, it typically assumes equal value for all unlabeled samples and randomly selects unlabeled data for manual annotation. This may lead to suboptimal use of annotation budgets. AL, by contrast, selects the most informative samples for manual labeling, often based on uncertainty, diversity, or representativeness [15]. Hybrid SSL_AL frameworks have gained attention for their ability to combine the strengths of both approaches.

Entropy-based uncertainty is one of the most widely used metrics for active sample selection. It identifies regions where the model is uncertain, thereby targeting ambiguous or hard-to-learn samples [27]. In 2022, Zhao *et al.* introduced BoostMIS [18], which combines entropy and model disagreement in a boosting framework to improve pseudo-label reliability while guiding the selection of high-value samples. More recently, Li *et al.* proposed HAL-IA [28], a hybrid framework that combines pixel-wise entropy, region-level consistency, and image-level diversity to guide interactive annotation. Both methods demonstrate the effectiveness of entropy-based scoring, especially when integrated with structural information or user feedback.

However, a key limitation of conventional AL strategies is their reliance on iterative annotation and training loops, where a model is repeatedly retrained after each round of human labeling. This process is labor-intensive and difficult to scale in clinical workflows where rapid deployment and annotation scalability are crucial.

To address these limitations, we introduce an entropy-aware semi-supervised segmentation framework that draws inspiration from AL but avoids iterative annotation loops. Rather than using entropy to select samples for manual labeling, we integrate entropy-based filtering directly into the training framework. Pseudo-labeled samples with low uncertainty are discarded, allowing the model to focus on uncertain samples. This approach retains the benefits of AL-driven sample prioritization while preserving the scalability and automation of end-to-end SSL training.

## III. METHOD

### A. Framework Overview

Our framework is divided into three components: SL part, SSL part, and an AL part. All components are built on top of nnU-Net's standardized framework (both 2D and 3D), benefiting from its consistent preprocessing and augmentation strategies.

### B. Semi-Supervised Learning Part

As shown in Figure 1, the SSL part is built on the standardized preprocessing and training framework of nnU-Net. Suppose we have two datasets: the labeled dataset $D\{X, Y\}$ and the unlabeled dataset $D_u\{X_u\}$. The training framework begins by planning and applying preprocessing steps on the labeled dataset. To ensure consistency between labeled and unlabeled samples, all unlabeled data are subjected to the same preprocessing transformations derived from the labeled dataset. This includes identical intensity normalization, voxel spacing resampling, and cropping strategies.

After this step, we apply the weak-to-strong augmentation strategy inspired by FixMatch [14]. In the original FixMatch framework, weak augmentation typically includes simple transformations like random flipping and cropping, while strong augmentation involves more aggressive operations such as rotation and noise blurring.

In our framework, the weak augmentation is directly inherited from nnU-Net's default augmentation framework to produce pseudo label $\{Y_u\}$, which includes operations such as random scaling, Gaussian noise, brightness and contrast adjustment, and spatial transformations. To generate strongly augmented inputs, we apply additional transformations: random rotations and combinations of rotations and flipping on top of the nnU-Net-augmented images. This two-stage augmentation strategy enables the model to learn consistent predictions across varying levels of perturbation, improving its robustness to data variability in semi-supervised training. It also decreases the risk of overfitting under the low quantity of labeled datasets during training.

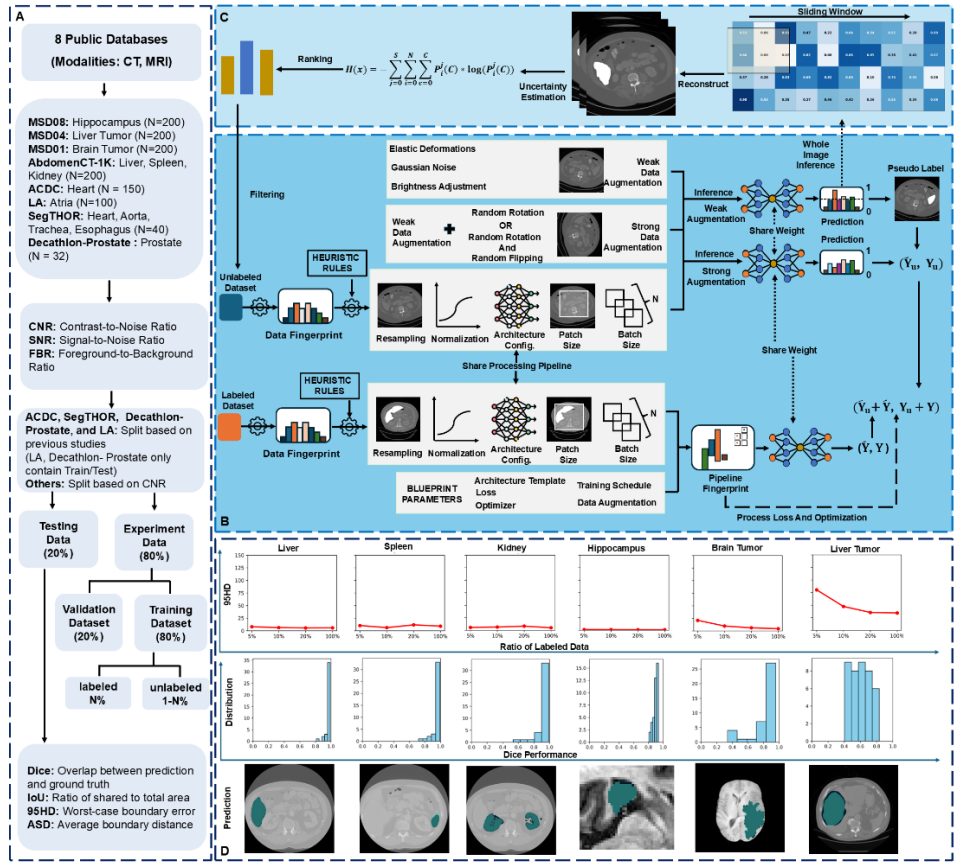

Fig. 1. Overview of the proposed framework. (A) Diagram illustrates the dataset preparation framework, including data analysis, preprocessing, splitting, and evaluation metric setup. (B) and (C) Diagram depict the streamlined integration of our SSL_AL approach into the standard nnU-Net framework. (D) Diagram presents the representative case study showcasing both qualitative and quantitative results across seven diverse segmentation tasks.

Another important component for the SSL is the calculation of the loss for prediction from both labeled and unlabeled datasets. Unlike traditional FixMatch, which introduces separate losses for labeled and unlabeled data, our method combines predictions from both labeled and unlabeled sources, and we compute a single unified loss using the nnU-Net's recommended loss function:

$$L = L\_seg \{(Y, Y\_u), (Y\_pred, Y\_u\_pred)\} \quad (1)$$

This design avoids complications arising from deep supervision conflicts and ensures stable joint optimization across labeled and pseudo-labeled samples.

Furthermore, instead of using the fixed confidence threshold of 0.95 suggested in the original FixMatch paper, we generate pseudo-labels by directly applying the SoftMax followed by an argmax operation at each pixel. In semantic segmentation tasks, every pixel is treated as an individual prediction point, leading to many predictions per image and often severe class imbalance. A static high threshold such as 0.95 may filter out a significant portion of useful predictions, especially from less confident or underrepresented regions such as tumor boundaries, small lesions, or minority tissue classes. By retaining all pixel-level predictions, including those with lower confidence, our approach ensures that uncertain but contextually informative regions contribute to training. This reduces the risk of reinforcing majority-class dominance and allows the model to better capture fine-grained structures. This change improves generalization from limited labeled data and enhances the overall effectiveness of pseudo-label supervision.

### C. Active Learning Part and Hybrid Approach

After a predefined number of training epochs, our AL mechanism is activated to selectively filter unlabeled data based on uncertainty estimation. The uncertainty strategy we used in our study is Shannon Entropy:

$$Entropy = -P(x)logP(x) \quad (2)$$

This uncertainty estimation is integrated into our SSL framework. This enhances the robustness of pseudo-labeling by prioritizing informative samples. Rather than following traditional AL paradigms that favor the inclusion of high-uncertainty samples for annotation, we adopt a reverse filtering approach: we discard a fixed percentage of low-uncertainty pseudo-labeled samples that are deemed redundant or uninformative based on the ranking of uncertainty for each unlabeled data.

This design is motivated by the observation that confident predictions made by the model on unlabeled data, especially when derived from itself, may not contribute significant new information to the learning process. Including these samples in training can lead to overfitting and reinforcing existing biases in the model. In segmentation tasks, where pixel-wise supervision is essential and class imbalance is common, blindly trusting confident predictions can suppress minority classes and reduce model generalizability.

By doing so, we maintain a balanced training signal and mitigate the risk of overfitting to dominant regions or classes.

**Algorithm 1: nnFilterMatch Learning Framework**

**Inputs:**
1, $D = \{(X, Y)\}$: labeled dataset
2, $D_u = \{(X_u)\}$: unlabeled dataset
3, $T_w(\cdot)$, $T_s(\cdot)$ weak and strong augmentation functions
4, $E_{warmup}$: warm-up epoch for supervised training
6, $E_{Maximum}$: max epoch for training
5, $E_{al}$: epoch to start active learning filtering
7, $f\theta(\cdot)$: segmentation model
8, $U(\cdot)$: uncertainty estimation function (e.g., entropy)
**Outputs:** Trained model $f\theta(\cdot)$

1: Initialize model $f\theta(\cdot)$ using nnU-Net backbone
2: for epoch E = 1 to $E_{Maximum}$
3:     if E ≤ $E_{warmup}$
4:         Train $f\theta(\cdot)$ on labeled data only:
5:             $L_{supervised} = Loss\ ([X], [Y])$
6:     elif > $E_{warmup}$
7:         Generate pseudo-labels for each $X_u \in D_u$
8:             Apply weak aug $T_w(X_u) \to X_w$
9:             Apply strong aug $T_s(X_w) \to X_s$
10:            Accept pseudo-label $Argmax(f\theta(X_w)) \to Y_u$
11:        Train use:
12:            $L = Loss\ ([f\theta(X), f\theta(X_s)], [Y, Y_u])$
13:    if E == $E_{al}$
14:        Estimate uncertainty $U(f\theta(X_u))$ for each data
15:        Remove n% low-uncertainty samples from $D_u$
16: end for
17: Return trained model $f\theta$

## IV. EXPERIMENT AND RESULT

### A. Dataset Detail

We evaluate our segmentation framework across diverse clinical tasks, involving 8 publicly available datasets that contain both tumor and organ segmentation.

**Automated Cardiac Diagnosis Challenge (ACDC)** [29]**:** The ACDC dataset consists of 150 cardiac cine-MRI exams across five categories: normal, myocardial infarction, dilated cardiomyopathy, hypertrophic cardiomyopathy, and abnormal right ventricle. Each exam includes manual segmentations of the left and right ventricles and myocardium at end-diastolic (ED) and end-systolic (ES) phases.

**AbdomenCT-1K** [31]**:** The AbdomenCT-1K dataset consists of 1,112 contrast-enhanced abdominal CT scans collected from 12 different medical centers, ensuring diversity in scanner vendors, imaging phases, and disease types. Each data contains four different organ segmentation tasks: liver, kidney, spleen, and pancreas. We use three organ segmentations from this dataset: liver, kidney, and spleen.

**MSD_HepaticVessel (08)** [32]**:** Medical Segmentation Decathlon task 08 targets the segmentation of hepatic vessels from contrast-enhanced abdominal CT scans and is extracted from 1 medical institution. The dataset includes 443 scans, each containing annotated blood vessels and live tumors. We use the liver tumor segmentation for this dataset in our study.

**MSD_BrainTumor (01)** [32]**:** Medical Segmentation Decathlon task 01 focuses on brain tumor segmentation using multi-modal MRI scans from 484 patients, sourced from the BraTS 2016 and 2017 challenges. The dataset is collected from 19 different institutions. Each case includes four imaging sequences: T1, T1-Gd, T2, and T2-FLAIR, allowing detailed analysis. The dataset provides expert-annotated labels for three tumor subregions—enhancing tumor, non-enhancing core, and peritumoral edema—encoded as classes 3, 2, and 1, respectively. In our study, we perform the brain whole tumor segmentation for this dataset.

**MSD_Hippocampus (04)** [32]**:** Medical Segmentation Decathlon task 04 focuses on the segmentation of the hippocampus from structural brain MRIs. The dataset includes 263 T1-weight MRI scans, all acquired and labeled at 1 Medical Center. Each scan includes manual annotations of the anterior and posterior hippocampus, assigned as labels 1 and 2, respectively. For this dataset, we perform the whole hippocampus segmentation in our study.

**MICCAI 2018 Left Atrium (LA) Dataset** [33]**:** The LA dataset contains 154 gadolinium-enhanced 3D MRI scans from patients with atrial fibrillation. Of these, 100 cases include expert manual annotations of the left atrial cavity and are publicly available for training, while the remaining 54 cases are reserved for evaluation in the official challenge. The dataset was primarily curated by the University of Utah's Center for Integrative Biomedical Computing and is widely used as a benchmark to evaluate 3D segmentation model performance.

**SegTHOR Dataset** [34]**:** The SegTHOR dataset (Segmentation of Thoracic Organs at Risk) provides 60 thoracic CT scans (40 training, 20 testing) with expert manual segmentations of heart, aorta, trachea, and esophagus. All scans derive from a single institution—the Centre Henri-Becquerel in Rouen, France—collected between February 2016 and June 2017.

**Decathlon Prostate** [32]: This contains 48 3D multi-parametric MRI scans acquired from multiple institutions, using T2-weight imaging and is collected from one single institution.

**Table I**
Image Analysis: CNR, SNR, FBR

| TASK | CNR | SNR | FBR (%) |
| --- | --- | --- | --- |
| ACDC | 1.09 | 2.59 | 3.90 |
| AbdomenCT-1K (liver) | 1.38 | 2.73 | 5.80 |
| AbdomenCT-1K (spleen) | 1.40 | 2.73 | 5.42 |
| AbdomenCT-1K (kidney) | 1.33 | 2.73 | 5.42 |
| MSD_HepaticVessel | 1.35 | 3.70 | 0.41 |
| MSD_BrainTumor | 3.15 | 4.98 | 1.20 |
| MSD_Hippocampus | 0.55 | 3.83 | 5.64 |
| LA | 1.93 | 7.25 | 0.70 |
| SegTHOR | 1.45 | -0.12 | 0.27 |
| Decathlon Prostate | 0.26 | 2.95 | 2.87 |

*B. Dataset Statistic and Analysis*

Table I compares image quality and segmentation complexity across ten benchmarks using three metrics: Contrast-to-Noise Ratio (CNR), Signal-to-Noise Ratio (SNR), and Foreground-to-Background Ratio (FBR). MSD_BrainTumor shows the highest CNR (3.15) and SNR (4.98). These values indicate strong tissue contrast and signal fidelity favoring accurate segmentation. In contrast, MSD_Hippocampus has a low CNR (0.55) despite a moderately high SNR (3.83), reflecting weak foreground and background contrast complicating model learning. ACDC exhibits low CNR (1.09), and Decathlon Prostate dataset shows the lowest CNR overall (0.26), suggesting challenging signal quality. AbdomenCT-1K (liver, spleen, kidney) presents moderate CNR values (1.38–1.40) and SNR around 2.73 but relatively high FBR (5.42–5.80%), indicating balanced pixel distribution. Conversely, MSD_HepaticVessel combines reasonable CNR (1.35) and SNR (3.70) with extremely low FBR (0.41%), pointing to severe class imbalance that could destabilize learning and limit generalization. Datasets such as LA (CNR 1.93, SNR 7.25) and SegTHOR (CNR 1.45, SNR -0.12) illustrate diversity in image quality: LA offers excellent signal clarity, while SegTHOR's negative SNR signals difficulties distinguishing structures. Overall, these results reveal substantial variability, such as tissue contrast, signal quality, and class balance, in our segmentation tasks.

*C. Implementation Detail and Result*

*1) Technical Setup*: All experiments are conducted using two computing environments. Locally, we use a workstation running Ubuntu (24.04) Linux, equipped with an Intel Core i9-14900HX CPU and an NVIDIA GeForce RTX 4090 GPU (16GB VRAM). Additionally, some experiments are processed on Google Colab Pro+, utilizing NVIDIA A100 GPUs (40GB VRAM) and Intel(R) Xeon(R) CPU. All models are implemented using the PyTorch deep learning framework.

*2) Evaluation Metrics*: To assess segmentation performance, we employ four widely used metrics: Dice Similarity Coefficient (Dice), Jaccard Index (IoU), 95th Percentile Hausdorff Distance (95HD), and Average Surface Distance (ASD). Dice and IoU measure the degree of overlap between predicted and ground truth masks, with values ranging from 0 to 100%, where higher values indicate better segmentation accuracy. Dice emphasizes agreement between sets, while IoU provides the stricter overlap measurement by penalizing mismatches more heavily. The 95HD evaluates the maximum surface distance while excluding extreme outliers, capturing the worst-case boundary deviation in a robust manner. ASD quantifies the average distance between the predicted and actual object boundaries, reflecting the overall boundary alignment. Lower values of 95HD and ASD correspond to more accurate boundary predictions. Together, these metrics provide a comprehensive evaluation of both regions overlap and are defined as:

$$Dice = \frac{2*TP}{FP+2*TP+FN} \qquad (3)$$

$$IoU = \frac{TP}{TP+FN+FP} \qquad (4)$$

$$95HD = max\{P95(d(a,B)), P95(d(a,A))\} \qquad (5)$$

$$ASD = \frac{1}{|A|+|B|}(\sum d(a,B) + \sum d(b+A)) \qquad (6)$$

*3) Training Strategy and Detail*: Since we use the nnU-Net preprocessing framework, nnU-Net controls training variables, such as model architecture, for each case. Beyond these variables, we do control variables, such as 300 training epochs and warmup epochs.

For our ACDC SOTA experiments, we adopt the same training, validation, and test splits as in prior studies to ensure a fair comparison. The ACDC setup includes 2 configurations: one with 5% labeled data (3 cases labeled, 67 unlabeled) and another with 10% labeled data (7 cases labeled, 63 unlabeled). Unlike previous studies that typically convert 3D volumes into 2D slices, we directly utilize a 2D patch size of 256×256 with a batch size of 12 for training. To account for the varying amounts of labeled data, we set the warm-up epochs to 5 for the 5% setup and 20 for the 10% setup. N% drop for AL was set to 15% for each 100 epoch. We use the same splitting strategy in study [26].

For our SegTHOR SOTA experiment, we follow the same training, validation, and test partitioning as reported in study [35]. The dataset comprises 40 CT scans and is preprocessed under two configurations: **(1)** 10% labeled data (3 labeled cases and 25 unlabeled cases) and **(2)** 20% labeled data (6 labeled cases and 22 unlabeled cases). Training is performed using a 2D patch size of 512×512 and a batch size of 12. All other experimental settings are identical to those used for the ACDC dataset.

For our SOTA experiment with Decathlon Prostate dataset, we use the 10% labeled dataset strategy (2 labeled cases and 20 unlabeled cases) as the same in study [36]. We apply the patch size of 320×320 with batch size of 12. Other experimental settings are identical to those used for the ACDC dataset.

The other 2D dataset training follows a slightly different setup to evaluate the generalizability of our model across diverse tasks. To simulate a low-annotation scenario, we sample 200 cases from each dataset. Each dataset is configured into four annotation settings: **5% labeled** (6 labeled, 134 unlabeled), **10% labeled** (14 labeled, 126 unlabeled), **20% labeled** (28 labeled, 112 unlabeled), **100% labeled** (140 labeled, 0 unlabeled). For each configuration, the SSL_AL experiments are directly compared with their SL counterparts using a fixed validation set of 20 cases and a test set of 40 cases. The warm-up epochs for the 5% and 10% labeled settings follow the same setup as in the ACDC task, while the 20% setting uses a warm-up of 20 epochs.

For the three-organ segmentation (liver, kidney, and spleen) tasks in AbdomenCT-1K, we use a patch size of 512×512 with a batch size of 6. For MSD_Hippocampus, the patch size is set to 56×56 with a batch size of 48. In the MSD_BrainTumor task, we use a patch size of 192×192 and a batch size of 30. For the liver tumor segmentation task, we use the patch size of 512×512 and a batch size of 6.

Beyond these 2D segmentation tasks, we also extend our SOTA experiment on the LA dataset with a 3D approach. For this task, we use the patch size of 16×256×256 with a batch size of 2. We use the same splitting strategy in study [26], which contains 80 data for training and 20 data for tests.

Table II

The SOTA performances across different models on the ACDC dataset. Higher values of Dice and IoU indicate better segmentation performance, whereas lower values of 95HD and ASD represent better boundary accuracy. (We directly use the model performance from the other study since we use the same splitting strategy on the dataset).

| Method | Ratio | | Evaluation Metrics | | | |
|---|---|---|---|---|---|---|
| | Labeled | Unlabeled | Dice (%) ↑ | IoU (%) ↑ | 95HD (voxel)↓ | ASD (voxel)↓ |
| URPC (2022) | 3 (5%) | 67 (95%) | 55.87 | 44.64 | 13.60 | 3.74 |
| SS-NET (2023) | | | 65.82 | 55.38 | 6.67 | 2.28 |
| DMD (2023) | | | 80.60 | 69.08 | 5.96 | 1.90 |
| UNIMATCH (2023) | | | 84.38 | 75.54 | 5.06 | 1.04 |
| URCA (2024) | | | 83.31 | - | 6.95 | 2.16 |
| CROSSMATCH (2024) | | | 88.27 | 80.17 | 1.53 | 0.46 |
| CGS (2025) | | | 88.83 | 80.62 | 2.42 | 0.66 |
| Ours (2025) | | | 89.56 | 81.63 | 2.39 | 0.63 |
| URPC (2022) | 7 (10%) | 63 (90%) | 83.10 | 72.41 | 4.84 | 1.53 |
| SS-NET (2023) | | | 86.78 | 77.67 | 6.07 | 1.40 |
| DMD (2023) | | | 87.52 | 78.62 | 4.81 | 1.60 |
| UNIMATCH (2023) | | | 88.08 | 80.10 | 2.09 | 0.45 |
| URCA (2024) | | | 87.86 | - | 4.21 | 1.36 |
| CROSSMATCH (2024) | | | 89.08 | 81.44 | 1.52 | 0.52 |
| CGS (2025) | | | 89.83 | 82.11 | 2.08 | 0.68 |
| Ours (2025) | | | 90.06 | 82.41 | 1.46 | 0.44 |

Table III

The SOTA performances across different models on the SegTHOR dataset. Higher values of Dice and IoU indicate better segmentation performance, whereas lower values of 95HD and ASD represent better boundary accuracy. (We directly use the model performance from the other study since we use the same splitting strategy on the dataset).

| Method | Ratio | | Evaluation Metrics | | | |
|---|---|---|---|---|---|---|
| | Labeled | Unlabeled | Dice (%) ↑ | IoU (%) ↑ | 95HD (voxel)↓ | ASD (voxel)↓ |
| SS-Net (2022) | **3** (10%) | **25** (90%) | 74.85 | 62.56 | 14.25 | 3.58 |
| CT-CT (2022) | | | 75.33 | 62.27 | 34.22 | 6.10 |
| DHC (2023) | | | 74.68 | 62.14 | 27.89 | 3.88 |
| BCP (2023) | | | 78.69 | 66.67 | 39.60 | 7.74 |
| CGS (2025) | | | 81.74 | 71.66 | 6.46 | 2.05 |
| Ours (2025) | | | 85.59 | 76.00 | 8.21 | 2.23 |
| SS-Net (2022) | **6** (20%) | **22** (80%) | 77.98 | 66.43 | 13.89 | 3.32 |
| CT-CT (2022) | | | 78.72 | 66.71 | 15.95 | 4.64 |
| DHC (2023) | | | 77.85 | 63.15 | 25.03 | 3.01 |
| BCP (2023) | | | 79.06 | 67.11 | 39.58 | 3.06 |
| CGS (2025) | | | 83.69 | 74.52 | 5.09 | 1.79 |
| Ours (2025) | | | 86.95 | 77.73 | 8.68 | 2.37 |

Table IV
The SOTA performances across different models on the Decathlon Prostate dataset. Higher values of Dice and IoU indicate better segmentation performance, whereas lower values of 95HD and ASD represent better boundary accuracy. (We directly use the model performance from the other study since we use the same splitting strategy on the dataset).

| Method | Ratio | | Evaluation Metrics | | | |
|---|---|---|---|---|---|---|
| | Labeled | Unlabeled | Dice (%) ↑ | IoU (%) ↑ | 95HD (mm)↓ | ASD (mm)↓ |
| ICT (2022) | **2** (10%) | **20** (90%) | 39.91 | 28.95 | 24.73 | 7.59 |
| MCNetV2 (2022) | | | 40.58 | 28.77 | 21.29 | 7.11 |
| INCL (2023) | | | 55.67 | 41.91 | 31.09 | 15.78 |
| DiffRect (2024) | | | 62.23 | 48.64 | 10.36 | 3.41 |
| Our (2025) | | | 71.62 | 59.87 | 6.69 | 1.68 |

Table V
Generalizability study across diverse segmentation Tasks. Higher values of Dice and IoU indicate better segmentation performance, whereas lower values of 95HD and ASD represent better boundary accuracy.

| Clinical Task | Ratio | | Evaluation Metrics | | | |
|---|---|---|---|---|---|---|
| | Labeled | Unlabeled | Dice (%) ↑ | IoU (%) ↑ | 95HD (voxel)↓ | ASD (voxel)↓ |
| Liver | **140** (100%) | **0** (0%) | 97.21 | 94.76 | 5.69 | 1.55 |
| Spleen | | | 96.33 | 93.21 | 9.14 | 2.26 |
| Kidney | | | 94.55 | 90.42 | 4.99 | 1.50 |
| Hippocampus | | | 90.06 | 82.03 | 1.13 | 0.41 |
| Brain Whole Tumor | | | 90.17 | 82.75 | 4.52 | 1.12 |
| Liver Tumor | | | 58.11 | 46.27 | 34.40 | 9.55 |
| Liver | **6** (5%) | **134** (95%) | 95.89 | 92.44 | 7.79 | 2.43 |
| Spleen | | | 94.24 | 90.03 | 10.31 | 2.70 |
| Kidney | | | 92.98 | 87.91 | 5.72 | 1.43 |
| Hippocampus | | | 87.29 | 77.63 | 1.50 | 0.53 |
| Brain Whole Tumor | | | 80.79 | 70.45 | 21.05 | 7.10 |
| Liver Tumor | | | 22.68 | 15.85 | 80.37 | 42.66 |
| Liver | | | 96.61 | 93.63 | 6.30 | 1.87 |
| Spleen | | | 94.74 | 90.81 | 6.15 | 1.74 |
| Kidney | **14** | **126** | 91.21 | 86.23 | 6.41 | 1.68 |
| Hippocampus | **(10%)** | **(90%)** | 88.84 | 80.10 | 1.28 | 0.50 |
| Brain Whole Tumor | | | 86.21 | 76.35 | 9.76 | 3.34 |
| Liver Tumor | | | 37.57 | 29.14 | 47.33 | 21.55 |
| Liver | | | 96.89 | 94.15 | 5.44 | 1.38 |
| Spleen | | | 95.28 | 91.49 | 11.61 | 3.81 |
| Kidney | **28** | **112** | 93.00 | 87.93 | 8.22 | 2.35 |
| Hippocampus | **(20%)** | **(80%)** | 89.53 | 81.19 | 1.18 | 0.44 |
| Brain Whole Tumor | | | 89.49 | 81.40 | 6.11 | 1.63 |
| Liver Tumor | | | 54.64 | 43.46 | 35.44 | 10.13 |

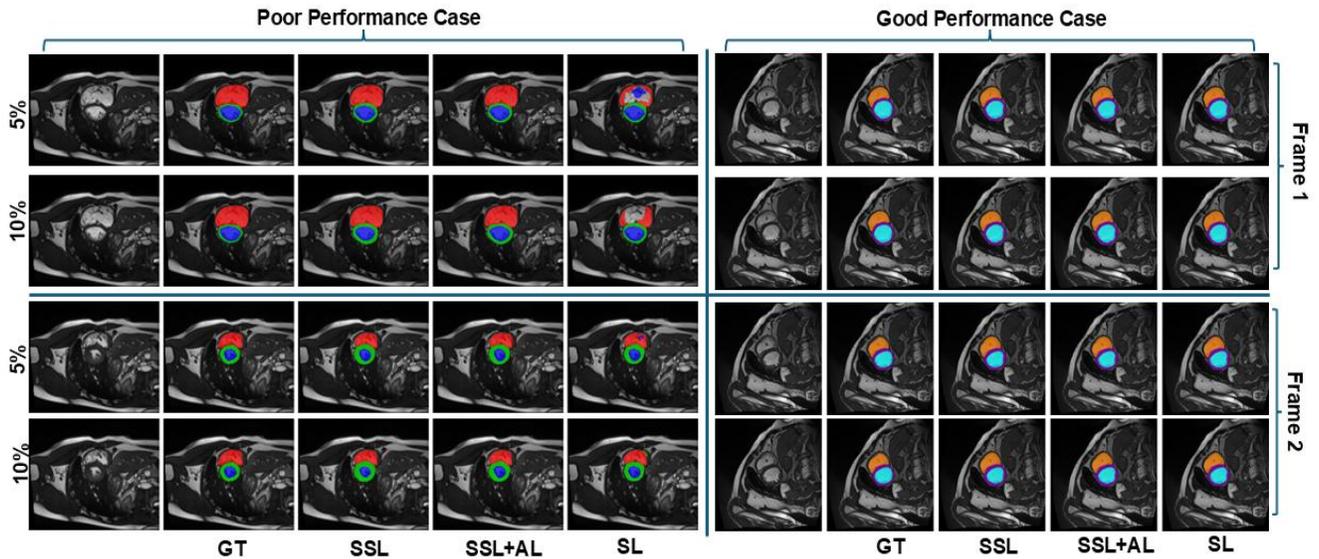

Fig. 2. Qualitative results from the ablation study on the ACDC dataset under 5% and 10% labeled conditions. Two representative cases are shown: one illustrating a case with suboptimal segmentation performance, and the other highlighting a case with high-quality predictions. Red, Blue, Green: Right Ventricle, Left Ventricle, and Myocardium for the poor performance case. Orange, Purple, Cyan: Right Ventricle, Left Ventricle, and Myocardium for the good performance case.

### D. The SOTA Comparison On 2D Cases

In this section, we discuss our model performance compared to other 2D SOTA models and the ablation study on ACDC, SegTHOR, and Decathlon Prostate dataset.

Table II summarizes our model's performance on the ACDC dataset. Compared to prior approaches, such as CrossMatch [26] and CGS [35], our method achieves higher Dice and IoU scores under both 5% and 10% labeled data settings. With 5% labeled data, our model achieves a Dice score exceeding 89%. With 10% labeled data, it surpasses 90%, demonstrating strong performance in low-annotation situation. Furthermore, our 95HD and ASD metrics reach SOTA levels, with values of 1.46 and 0.44 respectively under the 10% labeled setting.

Table III presents our model's performance on the SegTHOR dataset. Using 10% labeled and 90% unlabeled training data, our method achieves Dice and IoU scores of 85.59% and 76.00% respectively, which outperform recent 2025 methods such as CGS [35]. When increasing the labeled portion to 20% (80% unlabeled), the Dice and IoU further improve to 86.95% and 77.73%, respectively. However, the corresponding 95HD and ASD values are slightly worse than those reported by competing methods, indicating that while our model effectively segments the majority region of organ structures, it struggles to precisely capture fine anatomical boundaries and small irregular regions.

Table IV presents our model's performance on the Decathlon Prostate dataset. With 10% labeled data (2 labeled cases and 20 unlabeled cases), our method achieves a Dice score of 71.62% and an IoU of 59.87%, which is markedly higher than prior semi-supervised approaches such as DiffRect (62.23% Dice, 48.64% IoU) and INCL (55.67% Dice, 41.91% IoU). Furthermore, our 95HD (6.69 mm) and ASD (1.68 mm) are lower than those of competing methods, indicating more accurate boundary localization and smoother segmentation surfaces. These results confirm that even with limited labeled data, our approach generalizes well and delivers superior prostate segmentation quality compared to previous SOTA techniques.

## E. Ablation study on 2D SOTA Experiment

The ablation results presented in Table VI and VIII provide strong evidence for the effectiveness of our proposed SSL_AL framework. For the ablation study result of ACDC dataset, both SL and the SSL_AL approach consistently improved segmentation performance across all metrics and annotation levels compared to the fully SL baseline. Compared to the full SL baseline, both SSL and the SSL_AL approach consistently improve segmentation performance across all metrics and annotation levels. Notably, with only 5% labeled data, our SSL based framework improves the score by 6.55% and IoU performance by 8.58% compared to the pure nnU-Net framework. With the combined usage of our AL method, we push our segmentation framework to a better performance. At the 10% label ratio, the SL method improves the segmentation by 3.35% on dice score. However, our SSL method surpasses SL again by an additional 3.27%, reaching a Dice score of 89.58%. The integration of AL further pushes this boundary to 90.06%, highlighting the synergistic effect of combining SSL and AL in low-annotation regimes. Figure 2 shows the qualitative results on the ACDC dataset under two annotation ratios: 10% and 5%. We observe that the baseline SL struggles to capture complete structures, especially at lower annotation levels, missing key anatomical regions and producing fragmented outputs. In contrast, the SSL model recovers more complete shapes, while SSL_AL further enhances the segmentation accuracy, closely matching the ground truth.

Table VI
ACDC Ablation Study: SL vs. SSL vs. SSL_AL Performance Comparison

| Method | Labeled | Metrics | | | |
|---|---|---|---|---|---|
| | | Dice | IoU | 95HD | ASD |
| SL | 3 (5%) | 82.96 | 73.01 | 9.52 | 2.69 |
| SSL | | 89.51 | 81.59 | 2.43 | 0.71 |
| SSL_AL | | 89.56 | 81.63 | 2.39 | 0.63 |
| SL | 7 (10%) | 86.31 | 77.55 | 4.42 | 1.14 |
| SSL | | 89.58 | 81.65 | 2.41 | 0.69 |
| SSL_AL | | 90.06 | 82.41 | 1.46 | 0.44 |

Table VII
SegTHOR Ablation Study: SL vs. SSL vs. SSL_AL Performance Comparison

| Method | Labeled | Metrics | | | |
|---|---|---|---|---|---|
| | | Dice | IoU | 95HD | ASD |
| SL | 3 (10%) | 84.74 | 74.72 | 7.32 | 2.36 |
| SSL | | 85.25 | 75.53 | 8.74 | 2.38 |
| SSL_AL | | 85.59 | 76.00 | 8.21 | 2.23 |
| SL | 6 (20%) | 84.85 | 76.16 | 7.59 | 1.78 |
| SSL | | 86.81 | 77.55 | 8.33 | 2.15 |
| SSL_AL | | 86.95 | 77.73 | 8.68 | 2.37 |

The same pattern is also shown in the ablation study of the SegTHOR dataset. Compared to the full supervised training, the same pattern is also evident in the ablation study on the SegTHOR dataset. Compared to SL, both SSL and SSL_AL show higher Dice and IoU scores under both the 10% and 20% labeled settings. For example, under the 10% labeled setting (3 labeled subjects), Dice scores increase from 84.74% (SL) to 85.25% (SSL) and further to 85.59% (SSL_AL), with similar improvements observed in IoU scores (from 74.72% to 76.00%).

These improvements highlight the benefit of leveraging unlabeled data and active sample selection to enhance feature representation and boundary precision. Moreover, SSL_AL achieves the best overall performance, confirming that active learning further refines the segmentation quality beyond what semi-supervision alone can provide. While the Dice and IoU improvements are clear, the 95HD and ASD metrics show small variations in both datasets, suggesting marginally less consistent boundary localization.

## F. Cross Domain Generalization Performance

Beyond the 2D SOTA experiments, we further evaluate the generalizability of our model across diverse segmentation tasks.

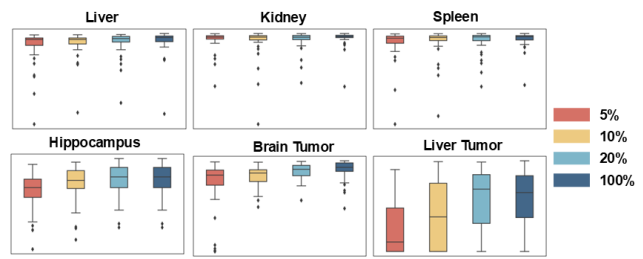

Fig.3. Box plot analysis across 6 segmentation tasks (Dice Score), highlighting performance variability and outliers.

1) Organ Segmentation Analysis: For high-contrast and well-structured organs such as the liver, spleen, and kidney. Our method achieves strong Dice scores even with minimal labeled data, as shown in Table V. At 10% annotation, we consistently observe that Dice scores exceed 90%, which closely match the fully supervised baseline. Hippocampus segmentation is another good sample that shows our model performance even with a low CNR, which is shown in Table I. With 10% annotation, we decrease the difference in dice sore performance between our proposed segmentation framework and default segmentation to 1.22%. These results confirm that our framework can accurately capture the core shape and volume of major anatomical structures with a fraction of the manual labels. While boundary-based metrics such as 95HD and ASD are slightly elevated for some cases with increased ratio, these differences are expected and might stem from the strong smoothing and deep supervision strategies employed by nnU-Net, which can produce softened boundaries. Nonetheless, the Dice and IoU performance demonstrates that the spatial overlap is largely preserved, making the results acceptable for those clinical tasks.

2) Tumor Segmentation Analysis: Tumor segmentation tasks, such as brain tumor segmentation, also demonstrate strong performance under low-annotation settings. The Dice performance gap between our proposed nnFilterMatch and the fully supervised nnU-Net baseline narrows significantly—from 9.38% at 5% annotation to only 0.68% at 20% annotation—highlighting the effectiveness of our method in minimizing annotation burden without sacrificing accuracy. Liver tumors present significantly greater challenges. At 20% labeled data, the Dice scores for liver tumors improve from 22.68% to 54.64%. The relatively low Dice scores observed in tumor segmentation tasks can be largely attributed to the intrinsic difficulty of these tasks. As shown in Figure 3, liver tumor segmentation performance improves more noticeably with increased annotation compared to other tasks, which exhibit greater fluctuation across different annotation ratios. Visualization analysis in Figure 4 reveals that these tumors occupy very small regions within the scans, making them more sensitive to annotation quantity.

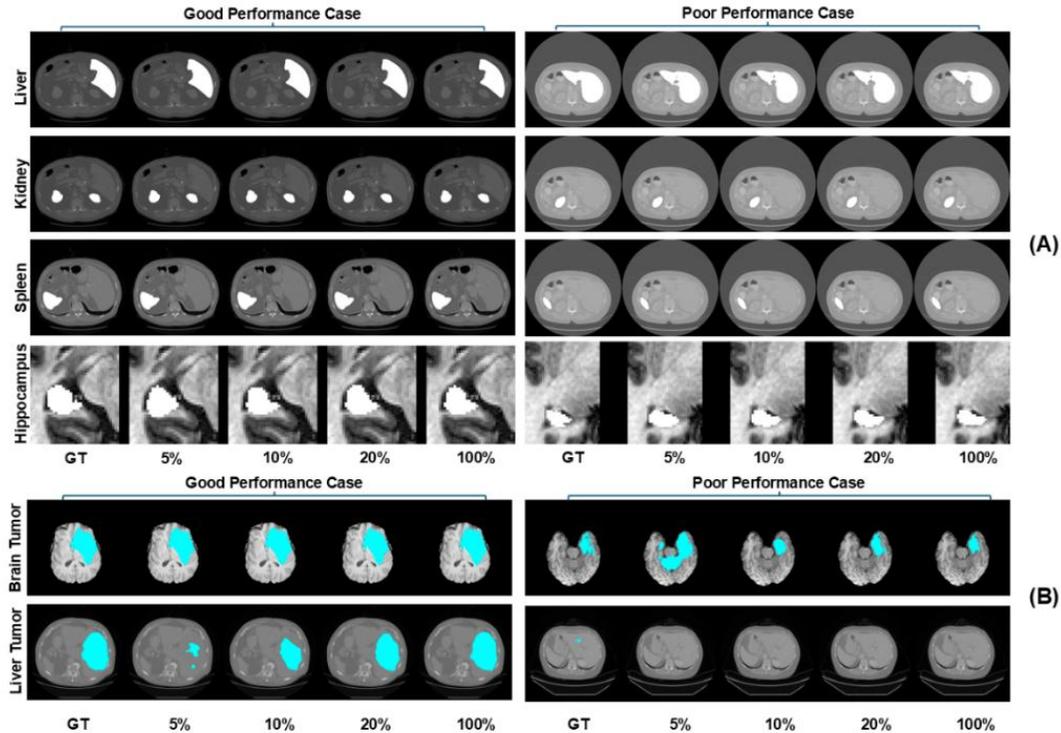

Fig. 4. Qualitative analysis of 6 clinical cases with both good performance and bad performance cases. (A) Figure shows the clinical tasks, including liver, spleen, kidney, and hippocampus. (B) Figure shows the tumor task, including brain whole tumor and liver tumor. White Mark: Organ Segmentation Task. Blue Mark: Tumor Segmentation Task.

Table VIII
The 3D model performance across different models on the LA dataset.

| Method | Ratio | | Evaluation Metrics | | | |
|---|---|---|---|---|---|---|
| | Labeled | Unlabeled | Dice (%) ↑ | IoU (%) ↑ | 95HD (voxel)↓ | ASD (voxel)↓ |
| SS-NET (2022) | **8 (10%)** | **72 (90%)** | 86.56 | 76.61 | 12.76 | 3.02 |
| MC-Net+ (2022) | | | 87.68 | 78.27 | 10.35 | 1.85 |
| DMD (2023) | | | 89.70 | 81.42 | 6.88 | 1.78 |
| BCP (2023) | | | 89.55 | 81.22 | 7.10 | 1.69 |
| UniMatch (2023) | | | 89.09 | 80.47 | 12.50 | 3.59 |
| CAML (2023) | | | 89.62 | 81.28 | 8.76 | 2.02 |
| RCPS (2024) | | | 90.73 | - | 7.91 | 2.05 |
| CROSSMATCH (2024) | | | 91.33 | 84.11 | 5.29 | 1.53 |
| Ours (2025) | | | 91.50 | 84.40 | 5.78 | 1.85 |
| SS-NET (2022) | **16 (20%)** | **64 (80%)** | 88.19 | 79.21 | 8.12 | 2.20 |
| MC-Net+ (2022) | | | 90.60 | 82.93 | 6.27 | 1.58 |
| DMD (2023) | | | 90.46 | 82.66 | 6.39 | 1.62 |
| BCP (2023) | | | 90.18 | 82.36 | 6.64 | 1.61 |
| UniMatch (2023) | | | 90.77 | 83.18 | 7.21 | 2.05 |

| Method | Labeled | Dice | Jaccard | 95HD | ASD |
|---|---|---|---|---|---|
| CAML (2023) | | 90.78 | 83.19 | 6.11 | 1.68 |
| RCPS (2024) | | 90.43 | - | 6.21 | 1.63 |
| BSNet (2024) | | 91.21 | - | 6.54 | 1.81 |
| CROSSMATCH (2024) | | 91.61 | 84.57 | 5.36 | 1.57 |
| Ours (2025) | | 91.79 | 84.90 | 5.58 | 1.83 |

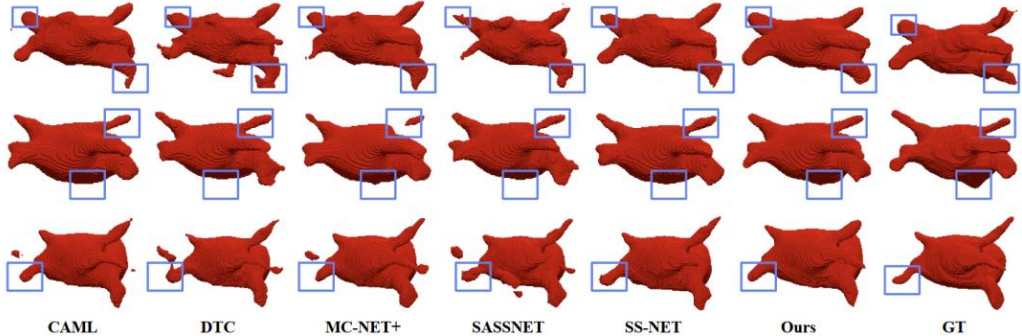

Fig. 5. Qualitative and visual analysis on three sample cases of LA dataset with different benchmark models with 8 labeled data and 72 unlabeled data.

### G. Experiment (SOTA) With 3D LA Dataset

Table IX
LA Ablation Study: SL vs. SSL vs. SSL_AL Performance Comparison

| Method | Labeled | Metrics | | | |
|---|---|---|---|---|---|
| | | Dice | IoU | 95HD | ASD |
| SL | 3 (10%) | 84.74 | 74.72 | 7.32 | 2.36 |
| SSL | | 85.25 | 75.53 | 8.74 | 2.38 |
| SSL_AL | | 85.59 | 76.00 | 8.21 | 2.23 |
| SL | 6 (20%) | 84.85 | 76.16 | 7.59 | 1.78 |
| SSL | | 86.81 | 77.55 | 8.33 | 2.15 |
| SSL_AL | | 86.95 | 77.73 | 8.68 | 2.37 |

We evaluate our proposed nnFilterMatch on the LA dataset using a full 3D setup to demonstrate its effectiveness in volumetric segmentation tasks. This experiment includes two different setups: 8 labeled and 72 unlabeled data, 16 labeled and 64 unlabeled data. Our nnFilterMatch also shows a powerful performance on the LA dataset with 3D segmentation, as shown in Table VIII. Compared to previous SOTA studies, our framework exceeds their performance. With 10% labeled and 90% unlabeled data, our framework achieves a Dice score of 91.50% and an IoU of 84.40%, surpassing the 2024 CrossMatch method, which reported 91.33% Dice and 84.11% IoU. Under the 20% labeled setting, our model further improves to 91.79% Dice and 84.90% IoU, consistently outperforming the prior SOTA. These results demonstrate not only the applicability of our method to 3D segmentation tasks but also its robustness under limited supervision.

The ablation study shown in Table IX also supports our model's performance in 3D cases. With 10% labeled data, both SSL and SSL_AL surpass the fully supervised baseline (SL) in Dice and IoU, demonstrating the benefit of leveraging unlabeled volumes. Increasing the labeled portion to 20% further boosts accuracy, with SSL_AL achieving the highest Dice (86.95%) and IoU (77.73%). Although the 95HD and ASD values fluctuate slightly across methods, the overall trend confirms the efficiency of SSL and that of AL together.

## V. Discussion

### A. Analysis of Framework

*1) Effect of nnU-Net*: nnU-Net plays a critical role in our study by providing an automatically adaptive framework for data preprocessing, architecture configuration, and training strategy planning, which ensures robust and consistent segmentation performance across diverse tasks. Its modular and task-aware design eliminates the need for manual tuning and allows seamless adaptation to different datasets and imaging modalities. With this framework, we significantly reduce the time and effort typically spent on model design, hyperparameter selection, and data normalization. This efficiency is particularly valuable in our semi-supervised and active learning setup, where experiments span multiple annotation ratios and clinical domains.

*2) Effect of Automated Model Configuration*: The self-adaptive model configuration of nnU-Net offers the ability to dynamically tailor network depth and architecture based on the characteristics of each dataset. This results in different structures being generated for different tasks. For example, the configuration for the 2D ACDC dataset is relatively shallow with feature map progression of (32, 64, 128, 256, 512, 512), whereas the 3D LA dataset—being more complex—uses a deeper configuration: (32, 64, 128, 256, 512, 512, 512, 512), enabling better feature extraction for larger anatomical variability.

*3) Effect of Initial Weight Decay and Learning Rate*: nnU-Net does not control the initial Learning Rate (LR) and Weight Decay (WD), but it does affect the model performance at certain ratio. As shown in Table X, the model gains a huge boom in performance when the LR and WD decrease from (0.01 and 0.00003) to (0.005 and 0.003).

**Table X**
ACDC 5% label performance with various LR and WD

| LR | WD | DICE | IOU |
|---|---|---|---|
| 0.01 | 0.00003 | 84.28 | 73.56 |
| 0.005 | 0.003 | 89.56 | 81.63 |

*4) Computational Resource Analysis*: We evaluate computational efficiency through a time cost analysis across six case studies, which compare our SSL framework to the SL baseline. Average epoch times for SL are reported for clarity. SSL generally requires higher computation per epoch, primarily due to additional forward passes and the inclusion of consistency loss for managing unlabeled data and generating pseudo-labels. An exception occurs in brain whole tumor task, where SL shows slightly higher time cost, likely from fully utilizing a densely annotated dataset. Overall, although SSL introduces extra computational overhead, the improved segmentation performance justifies the trade-off, highlighting the balance between efficiency and accuracy.

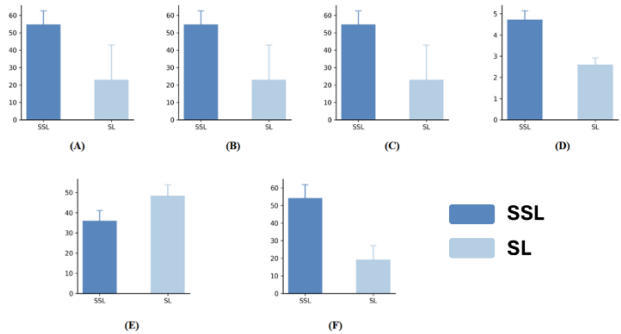

Fig. 6. Analysis for average epoch training time: A: Liver, B: Spleen, C: Kidney, D: Hippocampus, E: Brain Whole Tumor, F: Liver Tumor.

*5) Effect of SSL and AL*: Beyond nnU-Net, SSL and AL are two additional core components that strengthen our segmentation framework. Inspired by the FixMatch strategy [14], we integrate both strong and weak augmentation techniques to generate reliable pseudo-labels from unlabeled data. These pseudo-labels are then incorporated directly into the nnU-Net training framework, enabling the model to learn effectively from limited annotated data. While this step is relatively straightforward, it provides significant advantages. Based on the result from the ablation study in Table VI and the visualization result on Figure 2, our framework exceeds the pure nnU-Net framework on the ACDC (2D) segmentation task. Furthermore, results from Table VIII and Figure 5 for LA dataset (3D) provide another strong evidence to support the efficiency of our framework.

Our AL component is inspired by traditional uncertainty-based sampling strategies, particularly entropy-based selection and iterative annotation loops. While effective in theory, such strategies are often time-consuming and require frequent retraining, making them less suitable for real-time or large-scale applications. To overcome this limitation, we embed uncertainty-based selection directly into our end-to-end training framework. Instead of repeatedly processing the entire dataset for sample selection, we apply entropy-based filtering during training to dynamically exclude high-confidence pseudo-labeled samples. These focus on learning uncertain and informative cases, which not only enhances performance but also reduces the risk of overfitting to confidently misclassified unlabeled data. As shown in Table VI and Figure 2, the improvement in segmentation performance is consistent and achieves without any additional annotation effort on the ACDC dataset. Moreover, our framework retains compatibility with traditional active learning strategies, such as maximum entropy sampling, offering flexibility for future applications or hybrid designs.

*B. Limitation of Our Study*

*1) Limitation on High Challenge Tasks*: While our model demonstrates strong performance on several segmentation tasks, such as brain structure segmentation, it shows relatively weaker performance on highly challenging tasks, such as liver tumor segmentation. As illustrated in Table V and Figure 4, these tasks involve small, heterogeneous, or poorly defined regions, which are more susceptible to errors when using limited annotations and pseudo-labels. This highlights a potential limitation of our current framework in handling fine-grained or low-contrast targets, suggesting the need for further refinement or task-specific adaptation in future work. As shown in Figure 7, segmentation of liver tumors performance plateaus even with increased annotation ratios, indicating that simply adding large quantities of annotations does not guarantee stronger performance. This stagnation may reflect limitations in data quality, highlighting the need for improved curation or more sophisticated, task-specific learning strategies to achieve further performance improvements.

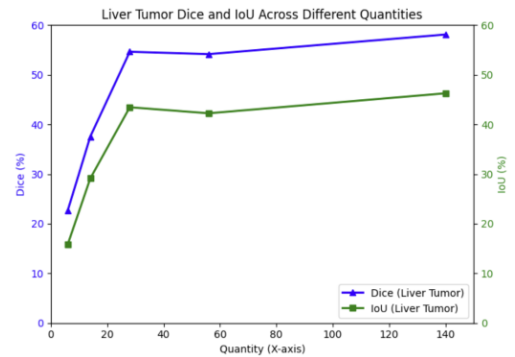

Fig. 7. Summary of segmentation performance on liver tumor datasets using 5%, 10%, 20%, 40%, and 100% labeled data.

*2) Other Limitation*: Our framework may have limitations in detecting fine-grained edges in small or subtle cases. Returning to the original nnU-Net study [19], the authors acknowledge that the model is primarily optimized for the Dice coefficient.

While this metric effectively captures volumetric overlap, it may bias the model against accurately capturing fine-grained boundaries or small structures, underscoring a potential limitation in tasks requiring high spatial precision.

Furthermore, nnU-Net is built upon the U-Net architecture, which is designed around a symmetric encoder–decoder structure with skip connections. This design captures both global context and fine-grained spatial information, but it can still struggle with segmenting small or low-contrast structures due to resolution loss in the encoder and potential feature [37, 38]. Furthermore, this structure might also limit the current extension on different algorithm, such as Transformer and Mamba.

Additionally, due to constraints in computational resources, we are unable to conduct an exhaustive set of experiments across all possible configurations. These aspects present opportunities for future investigation to further validate and enhance the robustness of our approach.

## VI. Conclusion

In this work, we present a novel end-to-end medical image segmentation pipeline nnFilterMatch that unifies the strengths of nnU-Net, semi-supervised learning and active learning within a single cohesive framework. Our approach leverages strong and weak augmentations in conjunction with entropy-based pseudo-label filtering to significantly reduce the need for manual annotation while maintaining high segmentation performance. Unlike traditional SSL_AL methods that rely on iterative retraining cycles, which are computationally expensive and often impractical for clinical integration, our method introduces a loop-free, single-pass training paradigm that is both scalable and clinically feasible.

Through extensive validation across multiple clinical segmentation tasks involving different organs, modalities, and difficulty levels, we demonstrate that the proposed method achieves performance comparable to or exceeding fully supervised models, even when using as little as 5–20% of labeled data. Our framework proves effective across diverse imaging domains including CT and MRI and consistently achieved robust results across tasks with varying anatomical complexity. These results suggest that our method can serve as a generalizable solution for segmentation in low-label regimes.

Importantly, beyond algorithmic improvements, this work contributes a practical advancement toward real-world deployment. The framework is designed to self-configure and requires minimal manual intervention or domain-specific tuning, making it accessible to non-expert users and adaptable across different clinical settings. By significantly lowering the annotation burden, it alleviates the demand on expert time and resources—two of the most significant bottlenecks in scaling medical AI.

Nevertheless, some limitations remain. Our entropy-based filtering approach, while effective overall, may suppress informative signals in highly complex cases or on small lesion targets. Future work will explore more adaptive uncertainty modeling techniques and refine pseudo-label selection mechanisms, possibly incorporating spatial priors or multi-scale entropy aggregation.

Looking forward to the future, we plan to extend our framework beyond segmentation to support downstream tasks such as diagnostic classification, lesion detection, and outcome prediction, forming a more comprehensive clinical AI pipeline. We also vision integrating temporal information for longitudinal studies and treatment monitoring, as well as exploring applications in low-resource or rare disease settings where annotation scarcity is particularly pronounced.

In summary, our proposed pipeline offers a robust, scalable, and annotation-efficient approach that significantly advances the practical deployment of medical AI systems. By substantially reducing reliance on fully annotated datasets while maintaining high clinical performance standards, this study sets a foundational precedent toward achieving truly integrated, end-to-end AI-driven clinical solutions.